\pdfoutput=1  % arxiv
\documentclass{article}

\usepackage{amsfonts}
\usepackage{microtype}
\usepackage{graphicx}
\usepackage{subfigure}
\usepackage{booktabs} % for professional tables

\usepackage{hyperref}
\usepackage[accepted]{icml2019}
\usepackage{amsmath}

\icmltitlerunning{Improving Neural Architecture Search Image Classifiers via Ensemble Learning}

\def\*{{\bf FIXME: }}
\DeclareMathOperator*{\argmin}{argmin}

\begin{document}

\twocolumn[
\icmltitle{Improving Neural Architecture Search Image Classifiers \\ via Ensemble Learning}



\icmlsetsymbol{equal}{*}

\begin{icmlauthorlist}
\icmlauthor{Vladimir Macko}{equal,google,resident}
\icmlauthor{Charles Weill}{equal,google}
\icmlauthor{Hanna Mazzawi}{google}
\icmlauthor{Javier Gonzalvo}{google}
\end{icmlauthorlist}

\icmlaffiliation{google}{Google Research, New York, NY, USA}
\icmlaffiliation{resident}{Work done as a member of the Google AI Residency program (\href{g.co/brainresidency}{g.co/brainresidency})}

\icmlcorrespondingauthor{Vladimir Macko}{vlejd@google.com}
\icmlcorrespondingauthor{Charles Weill}{weill@google.com}

\icmlkeywords{\*}

\vskip 0.3in
]



\printAffiliationsAndNotice{\icmlEqualContribution} % otherwise use the standard text.

{\begin{abstract}
Finding the best neural network architecture requires significant time, resources, and human expertise. 
These challenges are partially addressed by neural architecture search (NAS) which is able to find the best convolutional layer or cell that is then used as a building block for the network. 
However, once a good building block is found, manual design is still required to assemble the final architecture as a combination of multiple blocks under a predefined parameter budget constraint. 
A common solution is to stack these blocks into a single tower and adjust the width and depth to fill the parameter budget. 
However, these single tower architectures may not be optimal. 
Instead, in this paper we present the AdaNAS algorithm, that uses ensemble techniques to compose a neural network as an ensemble of smaller networks automatically. 
Additionally, we introduce a novel technique based on knowledge distillation to iteratively train the smaller networks using the previous ensemble as a teacher.  
Our experiments demonstrate that ensembles of networks improve accuracy upon a single neural network while keeping the same number of parameters. 
Our models achieve comparable results with the state-of-the-art on \texttt{CIFAR-10} and
sets a new state-of-the-art on \texttt{CIFAR-100}.

\end{abstract}

\section{Introduction}

% OUTLINE PROBLEM
Designing neural network (NN) architectures is often a demanding process. 
It often requires significant time, resources, and human expertise.
These challenges are partially addressed by neural architecture search (NAS), which is able to find the best convolutional layer or cell that is then used as a building block for the network ~\cite{real2018regularized, zoph2017learning}.
However, once a good building block is found, it is still required to manually design the final architecture as a combination of multiple blocks.
Moreover, there is usually a need to design multiple architectures with different parameter budgets, as different applications might pose different hardware constraints on memory and computation.

The critical question is how to upscale a small building block into a large architecture?
A common solution is to stack those blocks into a single tower and adjust the width and depth to fill the parameter budget. 
The solution of having one tower is common also for architectures that are not the result of neural architecture search \cite{springenberg2014striving,szegedy2015going,he2016deep}.

Recently, it was proposed to construct the network as an ensemble of smaller networks trained in a special way \cite{dutt2018coupled}.
Ensembles of neural networks are known to be much more robust and accurate than individual subnetworks.
Ensembles perform well on a wide variety of tasks \cite{caruana2004ensemble}
and are frequently used in the winning solutions of machine learning competitions (e.g. Kaggle) often consist of ensembles of multiple models. 
Unlike a single large neural network, an ensemble's size is not bounded by training, since each of the component subnetworks can be trained independently, and their outputs computed in parallel. 
Ultimately the final ensemble's size is bounded by its ability to fit on a serving hardware, and latency constraints.

% OUTLINE QUESTIONS
The main questions we tackle in this paper are the following: 
Can ensembles perform better than a single tower model with the same number of parameters? 
% When does knowledge distillation improves the performance?
Can we benefit from sequentially training the component subnetworks, and leverage information acquired from previously trained networks to improve the final ensemble performance?
Is it possible to construct ensemble architectures automatically or with minimal human expertise?
%

% OUTLINE PAPER SOLUTION/ contribution
In this work, we present a new paradigm to automatically generate ensembles of subnetworks that achieve high accuracy given a fixed parameter budget.
Our AdaNAS algorithm works in an iterative manner, and it increases the size of each new subnetwork at each iteration until the ensemble hits the budget limit.

As we iteratively learn the composition of the ensemble, we leverage information learned from subnetworks trained in previous iterations. 
We explore the effects of using ideas from Born Again Networks (BAN) \cite{furlanello2018born} to the final ensemble performance. 
In addition, we introduce a novel technique called Adaptive Knowledge Distillation (AKD) that extends Born Again Networks to use the previous iteration's ensemble as a teacher to assist in training the current iteration's subnetworks.

Resulting models are comprised of multiple separate towers that can be easily parallelized at inference time.
Our presented technique requires minimal hyperparameter tuning to achieve these results.
Our experiments demonstrate that ensembles of subnetworks improve accuracy upon a single neural network with the same number of parameters.
On \texttt{CIFAR-10} our algorithm achieves error $2.26$ and on \texttt{CIFAR-100} it achieves error $14.58$.
To our knowledge, and as we will show in Section~\ref{sec:results}, our technique achieves a new state-of-the-art on \texttt{CIFAR-100} compared to methods that do not use additional regularization or data augmentation (e.g., ShakeDrop~\cite{yamada2018shakedrop} or AutoAugment~\cite{cubuk2018autoaugment}).

This paper has been implemented as an extension of a framework for the construction and search of boosted ensembles, AdaNet~\cite{cortes2016adanet, weill2018adanet}.
The code to reproduce our results is available in the AdaNet project repository~\footnote{\href{github.com/tensorflow/adanet/tree/master/research/improve\_nas}{github.com/tensorflow/adanet/tree/master/research/improve\_nas}}. 
Our implementation uses open-sourced code provided by \cite{zoph2017learning} in the TensorFlow Models repository~\footnote{\href{github.com/tensorflow/models/tree/master/research/slim/nets/nasnet}{github.com/tensorflow/models/tree/master/research/\\slim/nets/nasnet}}.

This paper is organized as follows. We review previous work in Section~\ref{sec:related_work}. 
In Section~\ref{sec:ensembling_algorithm} we describe the ensembling algorithm. 
Experiment settings are outlined in Section~\ref{sec:experiment_setting} and Section~\ref{sec:results} shows the final results. Finally, Section~\ref{sec:discussion} discusses our proposed technique and our findings.

\section{Related work} %
\label{sec:related_work}

% TODO cite Genetic CNN, PNAS, ENAS, DARTS, SNAS, etc.

The work by~\cite{dutt2018coupled} showed a split into parallel branches. 
The main difference with previous ensembles was a tighter coupling of these branches
by placing an averaging layer before the softmax layer. 
These coupled ensembles were shown to outperform simple ensembles. 
In the best configuration of the coupled ensemble setting two models are trained without sharing any parameters. 

The main differences with our work are as follows. 
First, we train subnetworks sequentially. 
This allows us to use knowledge distillation techniques between iterations. 
Second, as we show in the following sections, one of our best configurations extends the average layer to become a weighted average of the different branches.
Due to this extension, each subnetwork is trained sequentially and a new weight is applied to each of them. 
Third, we demonstrate that we can achieve better accuracy with more than two subnetworks.

Regarding the neural network design part of our work, the work presented by~\cite{zoph2016neural, zoph2017learning} are the closest references. 
They search for an architectural building block on a small dataset (i.e., \texttt{CIFAR-10}) and then transfer the block to a larger dataset (e.g., \texttt{Imagenet}). 
Their main contribution is a new search space (i.e., NASNet) which produces a transferable cell architecture that improved upon the previous state-of-the-art. 
We use the cell found in the NASNet search space as our main building block in our experiments.

% knowledge distillation
Model Compression ~\cite{caruana2006compression} and Knowledge Distillation \cite{hinton2015distilling} are techniques for transferring the predictive power from a large ``teacher'' model or ensemble into a smaller ``student'' model. 
The goal of the student model is to learn the predictions of the teacher network, optionally in addition to the ground truth labels. 
Recent work by~\cite{furlanello2018born} has extended its application to be iterative, turning the student into the teacher at each iteration, and at the very end combining all the students's outputs to form an ensemble. 
Our work applies this technique, and extends it to use the ensemble of previous students as the teacher each iteration, instead of just the last student. 
Moreover, as mentioned earlier, we also explore ensembles with more than three students.

% ADANET
A related work that uses ensemble-learning for neural network design was presented by~\cite{cortes2016adanet}. 
Similarly, in this paper we use multiple subnetworks whose outputs are combined via a learned weighted average. 
However, unlike AdaNet we use subnetworks composed of stacked NASNet blocks. 
Furthermore, we improve candidate subnetworks training by using knowledge distillation techniques. 
Finally, the combination of the subnetworks is unconstrained and unregularized.

\section{AdaNAS algorithm} %
\label{sec:ensembling_algorithm}

    In this section we describe our ensembling algorithm.
    First, we describe it at a high level, and later, we present concrete realizations of each of its parts.

    \subsection{General algorithm}
    \begin{figure*}
    \begin{minipage}{.5\textwidth}
      \centerline{\includegraphics[width=\columnwidth]{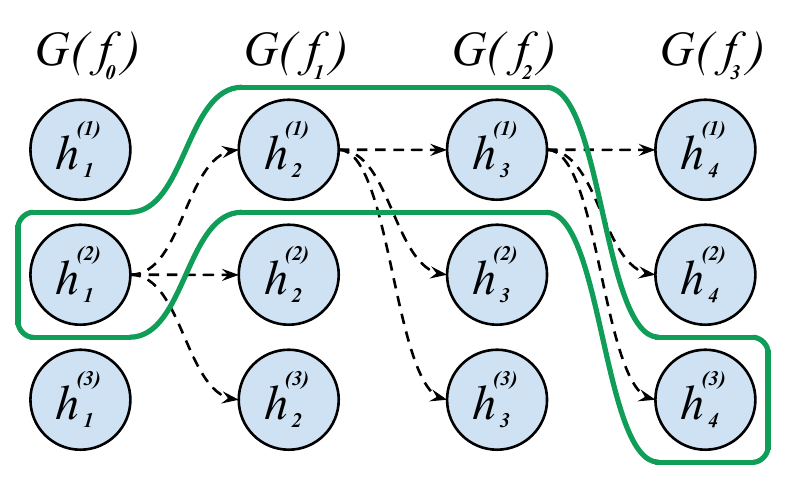}}
      \caption{
        Illustration of the search process over four iterations. 
        At each iteration, the search space is composed of three possible architectures to be explored ($h_i^{(j)}$, 
        for the $i$-th iteration and the $j$-th candidate, where $j \in [3]$). 
        The final ensemble is composed of the best subnetwork ($h_1^{(2)}, h_2^{(1)}, h_3^{(1)}, h_4^{(3)}$) from each iteration.}
      \label{fig:search}
    \end{minipage} \quad
    \begin{minipage}{.5\textwidth}
      \centerline{\includegraphics[width=\columnwidth]{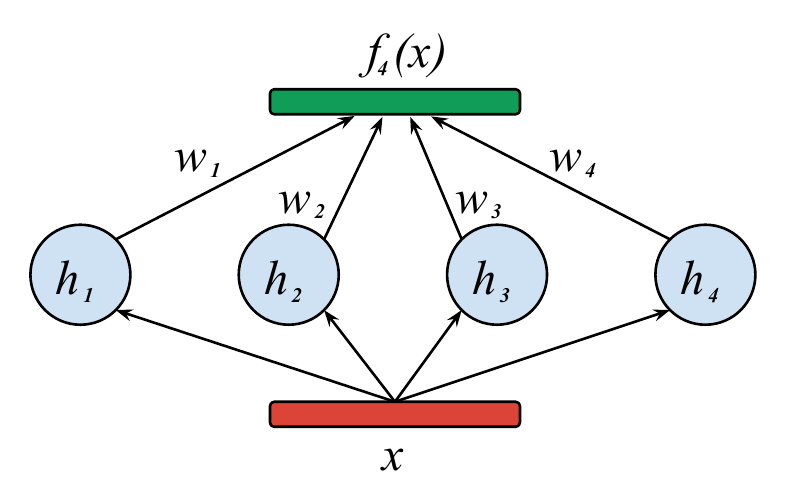}}
      \caption{
        Illustration of the final ensemble. 
        The output of each subnetwork $h_i$ is scaled by mixture weight $w_i$ and summed to form the final ensemble.}
      \label{fig:ensemble}
    \end{minipage}
    \end{figure*}

    We consider the standard supervised learning scenario and assume that training and test examples are drawn i.i.d. according to some distribution and denote by $S=\{(x_1,y_1) \ldots (x_m,y_m)\} \subseteq \mathcal{X} \times \mathcal{Y}$ a training set of size~$m$. 
    Additionally, for $I\in \mathbb{Z}$, we denote by $[I]$ the set~$\{1, 2, \ldots, I\}$.
    
    Our algorithm is iterative. 
    It starts with an empty ensemble~$f_0$. 
    In $I$ iterations it generates $I$ trained subnetworks $h_1,h_2,\ldots, h_{I}$ that will form an ensemble $f_I$.
    In each iteration $i\in [I]$ it proposes a set of candidate subnetworks called \textit{candidate subnetworks} $H_i = \{h_i^{(1)},h_i^{(2)}, \ldots\}$  (note that at this point, the candidate subnetworks are not trained and solely represent an architecture).
    By $h_i^{(j)}$ we denote the $j$-th candidate subnetwork in generated in iteration $i$.
    Generator $G$ is responsible for proposes candidate subnetworks based on the ensemble from the previous iteration, that is,~$H_i = G(f_{i-1})$.
    
    We denote by $p_{i,j}$ the set of parameters of the candidate subnetwork $h^{(j)}_i$. 
    Given a set of real numbers $\theta \in \mathbb{R}^{|p_{i,j}|}$. 
    We denote by $h_i^{(j)}(\theta)$ the resulting function when fixing the candidate subnetwork's parameters to $\theta$.
    
    The algorithm trains the parameters $p_{i,j}$ of each candidate subnetwork $h_i^{(j)}$ independently. 
    When training, the algorithm tries to find a $\theta \in \mathbb{R}^{|p_{i,j}|}$ that minimizes a classification loss $\mathcal{L}_h(S, h_i^{(j)}(\theta))$ as well as a knowledge distillation loss $\mathcal{L}_{KD}(S, f_{i-1}, h_i^{(j)}(\theta))$ against the previous ensemble~$f_{i-1}$. 
    
    The previous ensemble and its component subnetworks are kept frozen, and are not affected by training the candidate subnetworks. 
    By sharing a common computation subgraph, we can more efficiently train several candidate subnetworks within a single GPU.
    
    Once all candidate subnetworks are trained, the algorithm tries to add each one to the previously constructed ensemble $f_{i-1}$; It chooses the one that minimizes the loss of the whole ensemble.
    The best candidates subnetwork $h_i^{(*)}$ then becomes a part of the ensemble as a subnetwork $h_i$.
    The logits of the ensemble $f_i$ are a weighted sum of the trained subnetworks' logits, that is,
    \begin{equation}
      f_i = \sum_{k=1}^i w_k \cdot h_k,
    \end{equation}
    where $w_{k}\in \mathbb{R}$ for $k\in[i]$, denotes the mixture weight and $h_k$ is the subnetwork in the $k$-th iteration, respectively. 
    Without loss of generality, we denote by $h_k$ the logits of the trained subnetwork selected in the $k$-th iteration.
    Note that the mixture weights are applied to each subnetwork's outputs before the last non-linearity, such as a softmax activation in the case of multi-class classification.

    In the experiments section we also explore a simpler more restricted version of the algorithm where the logits of the ensemble is just an average of previously selected subnetworks' logits, that is, $f_i = \frac{1}{i}\Sigma_{k=1}^i h_k$.
    
    The vectors of mixture weights $w^{(k)}$ is trained to minimize the classification loss of the final ensemble. 
    During training, the algorithm alternates between solving two optimization problems: training the mixture weights $w^{(k)}$ and training the candidate subnetworks $h_i^{(k)}$.
    
    Once the $i$-th subnetwork is selected by the algorithm, it advances to the next iteration. 
    After $I$ iterations, the algorithm outputs the final ensemble $f_I$ of the last iteration.

    \begin{algorithm}[tb]
    \caption{AdaNAS algorithm}
    \label{alg:example}
    \begin{algorithmic}
       \STATE {\bfseries Input:} $S$, $I$, $G$
       \STATE $f_0 \gets 0$
       \FOR{$i=1$ {\bfseries to} $I$}
           \STATE $H_i \gets G(f_{i-1})$
                \FOR{$h_i^{(j)}$ {\bfseries in} $H_i$}
                    \STATE $\theta^{(j)} \gets \argmin_{\theta} 
                            \mathcal{L}_h(S, h_i^{(j)}(\theta)) +$
                    \STATE ~~~~~~~~~~ ~~~~~~~~~~ ~~~~~ $\mathcal{L}_{KD}(S, f_{i-1}, h_i^{(j)}(\theta))$
                    \STATE $w^{(j)} \gets \argmin_w \mathcal{L}_f(S, \Sigma_{k=1}^{i-1}w_{k}^{(j)} \cdot h_k + w_{i}^{(j)} \cdot h_i^{(j)})$
                \ENDFOR
           \STATE $j^* \gets \argmin_{j}
           \mathcal{L}_f(S, \Sigma_{k=1}^{i-1}w_k^{(j)} \cdot h_k + w_i^{(j)} \cdot h_i^{(j)}(\theta^{(j)})) $
           \STATE $h_i \gets h_i^{(j^*)}(\theta^{(j^*)}), w \gets w^{(j^*)}$
           \STATE $f_i \gets \Sigma_{k=1}^i w_k h_k$
       \ENDFOR
       \STATE \textbf{return} $f_I$
    \end{algorithmic}
    \end{algorithm}

    After presenting the general algorithm, we describe concrete possible realizations by discussing possible generators~$G$,
    knowledge distillation losses $\mathcal{L}_{KD}$, and mixture weight learning procedures we have used to train the ensemble.

    \subsection{Algorithm variants}

    For classification loss $\mathcal{L}_h$ we use cross entropy $\mathcal{H}$.
    For knowledge distillation loss $\mathcal{L}_{KD}$ we propose three options. 
    We can use previous subnetwork as the teacher (BAN)
    \begin{equation}
        \mathcal{L}_{BAN}(S, f_{i-1}, h_i^{(j)}(\theta)) = \mathcal{H}(h_{i-1}, h_i^{(j)}(\theta))
    \end{equation}
    
    We also propose a novel loss based on knowledge distillation \cite{hinton2015distilling}
    that instead uses the previous iteration's entire ensemble as the teacher, hence adaptive knowledge distillation (AKD):
    \begin{equation}
        \mathcal{L}_{AKD}(S, f_{i-1}, h_i^{(j)}(\theta)) =  \mathcal{H}(f_{i-1}, h_i^{(j)}(\theta))
    \end{equation}
    
    Finally, we can turn knowledge distillation off completely (NOKD) $\mathcal{L}_{NOKD} = 0$.

    We train mixtures weights $\mathbf{w}$ concurrently yet independently from the candidate subnetwork parameters. 
    The gradients from the mixture weights do not propagate through the subnetworks, and therefore do not affect subnetwork training. 
    We do this efficiently by using different train ops in TensorFlow for each subnetwork.
    We also explore a case where mixture weights are not trained and just set to uniform values $w_i^{(k)} = \frac{1}{i}$.
    
    For generator $G$ we discuss two main variants.
    $G_C$, the ``constant'' generator, always generates a single candidate with the same architecture. 
    This subnetwork is always added to the ensemble at the end of the iteration.
    
    In general, generator $G$ can decide what architectures to propose based on the previously selected candidates (based on the architectures of subnetworks already present in the ensemble).
    We propose a dynamic generator $G_D$.
    $G_D$ at iteration $i$ looks at the previously selected subnetwork $h_{i-1}$ and  generate one candidate that is deeper (more layers/blocks/cells), and one that is wider (more convolution channels).
    
    Even though generators can propose very different architectures, we train them with the same hyperparameters regardless. 
    This could be addressed by generator proposing hyperparameters as well as architectures. 
    However this approach risks a combinatorial explosion of candidates.
    
    As an illustration, if we use the constant generator ($G_C$), no knowledge transfer ($\mathcal{L}_{NOKD}=0$), do not train $w_i$ (set uniformly to $\frac{1}{I}$),
    this algorithm is equivalent to a uniform-average ensemble of $I$ independently trained subnetworks.

\section{Experimental setting}
\label{sec:experiment_setting}

    In this section we describe the experimental conditions used to obtain the results shown in Section~\ref{sec:results}. 
    First, we fix our parameter budget to 33M parameters similar to the larger NASNet-A(7@2304) as this is the main baseline we try to improve upon.

    In notation NASNet-A(X@Y), $X$ describes the depth of the model (in terms of cells) and $Y$ describes the number of convolution channels.

    We explore two sets of experiments. The first uses a constant generator combined with different knowledge distillation methods to learn the ensemble. 
    The second, explores a restricted search space of candidate subnetwork architectures, greedily growing the ensemble with the subnetwork that most improves its performance.
    
    For all our experiments we use the same hyperparameters as described in~\cite{zoph2017learning} with no extra tuning.
    We use batch size $32$, learning rate $0.025$, cosine learning rate schedule, momentum optimizer with momentum $0.9$ and we clip gradients to $5$.
    We also use $1M$ training steps per iteration.

    All our experiments are run on 11 NVIDIA Tesla V100 GPUs with asynchronous gradient descent employing data parallelism.
    
    \subsection{Data}
    We ran all experiments on two well calibrated datasets, \texttt{CIFAR-10} and \texttt{CIFAR-100}~\cite{krizhevsky2009learning}.
    Both datasets consists of $60,000$ RGB images uniformly spread across $10$ and $100$ classes, respectively. 
    Images are preprocessed in the following way.
    All images are upscaled to size $40\times40$, randomly croped back to $32\times32$ and randomly horizontally flipped. 
    Finally the images are whitened.
    This procedure follows \cite{zoph2017learning} as other works. 
    We additionally used Cutout as the augmentation technique~\cite{devries2017improved}.
    
    \subsection{Constant generator}
    
    The first set of experiments focuses on the case where a constant generator $G_C$ is used. 
    This generator always proposes a single candidate subnetwork. 
    In our experiments, the architecture of this candidate will be NASNet-A(6@768) $h_{6@768}$, hence $G_C(f_i)= H_i = \{h_{6@768}\}$.
    Once the candidate subnetwork is trained it is added directly to the ensemble as it is the only candidate subnetwork.
    
    Three types of experiments are analyzed depending on the knowledge distillation technique used: Adaptive Knowledge Distillation (AKD), Born Again Networks (BAN) and no knowledge distillation (NOKD). 
    Mixture weights are learned for AKD and NOKD. 
    Note that we have intentionally left BAN out of this option as we observed that BAN performs worse without mixture weights.
    
    The algorithm in these experiments has a maximum of $10$ iterations, which generates ensembles of $10$ NASNet-A(6@768) models. 

    \subsection{Growing architecture}
    The objective of the second set of experiments is to explore two dynamic generators. 
    The first generator $G_{D6@768}$ proposes two types of candidates, a deeper one, (NASNet-A(7@768) in the first iteration), and a wider one, (NASNet-A(6@1008) in the first iteration). 
    During the first iteration it uses NASNet-A(6@768) as a starting point. 
    During the following iterations it uses the previously chosen subnetwork to generates one candidate that is deeper by one cell and one that is wider by $240$ convolution filters. 
    The number $240$ was chosen arbitrarily as a reasonable small constant.

    The second generator $G_{D1@240}$ which is identical to the previous generator but the starting point is a minimal architecture NASNet-A(1@240).
    This generator has less human bias regarding the model architecture.
    
    Despite training multiple candidate subnetworks at each iteration, we only select the best subnetwork to add to the ensemble. 
    We qualify as ``best'' the candidate subnetwork that most improves the ensemble's performance on the full training set, thereby eliminating the need for a hold-out set.

\section{Results}
\label{sec:results}

    \subsection{Baselines}
    
    For baselines we do not consider regularization techniques such as Shake-shake~\cite{gastaldi2017shake} or Shake-drop~\cite{yamada2018shakedrop} nor novel data augmentation techniques like AutoAugment~\cite{cubuk2018autoaugment}.
    
    The main baseline we try to improve upon is NASNet-A(7@2304) which reaches an error of $2.40$ on \texttt{CIFAR-10}.  
    We managed to reproduce these results to $2.41 \pm 0.04$ ($2.47 \pm 0.06$ without early stopping).

    Throughout the experiments we will use NASNet-A(6@768) that should achieve error $2.65$. 
    With hyperparameters provided in the paper we were able to achieve error $2.78 \pm 0.07$ on a single GPU ($2.82 \pm 0.07$ without early stopping).
    This error slightly increased for training on 11 GPUs to $2.81 \pm 0.06$.
    This is the starting point for most of our algorithms, as this is the error achieved at the end of the first iteration.
    See supplement for thorough breakdown of hyperparameters that we used.

    As most of our experiments are run for $10M$ steps with 11GPUs, we benchmark NASNet-A(7@2304) in the same conditions:
    This substantially increased the error to $3.54 \pm 0.11$ ($3.69 \pm 0.17$ without early stopping).

    On \texttt{CIFAR-100} NASNet-A(6@768) reportedly achieves error $16.58$ and NASNet-A(7@2304) $16.03$ \cite{luo2018neural,zoph2016neural}.

    We used the same hyperparameters and achieved error $17.70 \pm 0.28$ ($17.87 \pm 0.29$) for NASNet-A(7@2304) and $17.58 \pm 0.19$ ($17.71 \pm 0.13$) for NASNet-A(6@768).

    \subsection{Ensembling}

    As seen in Table~\ref{tab:ourresults}, we were able to improve upon our NASNet-A(7@2304) baseline.
    By ensembling $10$ small models we achieved relative $5\%$ reduction in error (absolute $0.11$) on \texttt{CIFAR-10} and relative $9\%$ reduction in error (absolute $1.45$) on \texttt{CIFAR-100}.
    We observe that with uniform mixture weights it is best to not use any knowledge distillation (NOKD). 

    This is a positive result for practical applications, as it means that different subnetworks do not need to be trained sequentially, and instead could be trained in parallel. 
    
    Note, that the NASNet baseline on \texttt{CIFAR-100} was probably trained with different parameters than the ones used on \texttt{CIFAR-10}. 
    As our reproduction suggests, hyperparameters from \texttt{CIFAR-10} do not translate well to the \texttt{CIFAR-100}.
    However, our algorithm achieved state-of-the-art results even though the hyperparameters were not fine tuned on this dataset.

\begin{table}[t]
    \caption{Accuracy error for different types of knowledge distillation applied to the candidates in the ensemble. All models were trained with cutout for a fixed number of steps without early stopping. Different types of knowledge distillation are denoted as NOKD (no knowledge distillation), AKD (adaptive knowledge distillation) and BAN (born again knowledge distillation). Experiments with mixture weight training are denoted with $w$.}
    \label{tab:ourresults}
    \vskip 0.15in
    \begin{center} \begin{small} \begin{sc}
        \begin{tabular}{lcc}
            \toprule
            Type of KD & \texttt{CIFAR-10} & \texttt{CIFAR-100} \\
            \midrule
NOKD    & $2.29 \pm 0.07$  & $\mathbf{14.58 \pm 0.05}$ \\
AKD     & $2.30 \pm 0.03$  & $14.94 \pm 0.16$ \\
BAN     & $2.37 \pm 0.04$  & $15.03 \pm 0.08$ \\
$w$ + NOKD    & $2.35 \pm 0.03$  & $14.87 \pm 0.19$ \\
$w$ + AKD     & $\mathbf{2.26 \pm 0.05}$  & $14.83 \pm 0.14$ \\
$G_{D6@768}$ + NOKD   & $3.23 \pm 0.02$  & $16.10 \pm 0.05$ \\
$G_{D1@240}$ + NOKD   & $3.06 \pm 0.04$  & $15.55 \pm 0.07$ \\
$G_{D6@768}$ + AKD    & $3.09 \pm 0.07$  & $16.96 \pm 0.10$ \\
$G_{D1@240}$ + AKD    & $3.10 \pm 0.11$  & $15.31 \pm 0.04$ \\
        \bottomrule
        \end{tabular}
    \end{sc}\end{small}\end{center}
    \vskip -0.1in
\end{table}

    We observe that by training mixture weights ($w+$) we were able to improve upon these results even further. 
    Training mixture weights increases the error for NOKD and decreases the error for AKD.
    On \texttt{CIFAR-10} we achieve error $2.26 \pm 0.05$ which is a $6\%$ relative improvement ($0.14$ absolute) upon NASNet baseline, which was a previous state-of-the art result.
    To our knowledge, this result is surpassed only by 
    AmoebaNet-B (6, 128) $2.13 \pm 0.04$ \cite{real2018regularized}, 
    but uses a different architecture search space, and NAONet(f=128) $2.11$ that has $4$ times more parameters ($128M$)  \cite{luo2018neural}.
    We do not consider results from methods that use additional regularization or data augmentation techniques.
    
    Using early stopping, these results can be further improved by approximately ~$0.1$.
    Our best run achieved error~$2.09$.
    
    On \texttt{CIFAR-10}, the performance of NOKD is on par with the large NASNet-A(6@768) baseline. 
    However, on \texttt{CIFAR-100} NOKD significantly outperforms the baseline.
    Learning mixture weights can improve performance in some cases. 
    In particular, AKD benefits from learning them more than NOKD. 
    We conclude that on \texttt{CIFAR-10} the best combination is to use Adaptive Knowledge Distillation and learn mixture weights. 
    On \texttt{CIFAR-100} it is best to use a uniform average ensemble of independently-trained models.
    
    We see, that even experiments with an adaptive search $G_{D1@240}$ can get similar results to a hand-crafted configuration on \texttt{CIFAR-100}.
    For \texttt{CIFAR-10}, these results are substantially worse the NASNet baseline.
    We observe, that tweaking the generator $G_{D1@240}$ to also reconsider the previous iteration's best architecture can achieve $2.38 \pm 0.06$ error on \texttt{CIFAR-10}, which is competitive with the NASNet baseline but requires less human expertise.

    $G_{D6@768}$ performs worse than or similar to $G_{D1@240}$. 
    One possible reason is that $G_{D6@768}$ only trains for $4$ iterations.  
    At the $5$-th iteration the model cannot fit the two proposed candidates into the V100 GPU's memory.

    \subsection{Comparison with other results}

    To the best to our knowledge, Table \ref{tab:cif10:others} contains current state-of-the-art results for \texttt{CIFAR-10}  and Table \ref{tab:cif100:others} for \texttt{CIFAR-100}.

    {

\begin{table}[t]
    \caption{Classification accuracy errors on \texttt{CIFAR-10}. Baseline results in this table are gathered from \cite{zoph2017learning,luo2018neural,real2018regularized}. Our approach (AdaNAS) uses adaptive knowledge distillation and mixture weights learning. This corresponds to $w$ + AKD from table \ref{tab:ourresults}.}
    \label{tab:cif10:others}
    \vskip 0.15in
    \begin{center} \begin{small} \begin{sc}
        \begin{tabular}{lcc}
            \toprule
            Model & \#parameters & error (\%) \\
            \midrule
            NASNet-A(6@768)         &  3.3M & 2.65  \\ % zoph2017learning 
            NAONet (f=36)           & 10.6M & 3.18  \\ % luo2018neural
            NAONet (f=64)           & 28.6M & 2.98  \\ % luo2018neural
            NASNet-A(7@2304)        & 32.6M & 2.40  \\ % zoph2017learning
            AmoebaNet-B (6, 128)    & 34.9M & $2.13 \pm 0.04$ \\ %real2018regularized
            NAONet (f=128)          &  128M & $2.11$  \\ % luo2018neural
            \midrule
            AdaNAS                  &   33M & $\mathbf{2.26 \pm 0.05}$ \\
            \bottomrule
        \end{tabular}
    \end{sc}\end{small}\end{center}
    \vskip -0.1in
\end{table}

}
    {

\begin{table}[t]
    \caption{Classification accuracy errors on \texttt{CIFAR-100}. Baseline results in this table are gathered from \cite{luo2018neural,zoph2016neural}. Our approach (AdaNAS) does not use knowledge distillation. This corresponds to NOKD from table \ref{tab:ourresults}.}
    \label{tab:cif100:others}
    \vskip 0.15in
    \begin{center} \begin{small} \begin{sc}
        \begin{tabular}{lcc}
            \toprule
            Model & \#parameters & error (\%) \\
            \midrule
            NASNet-A(6@768)      & 3.3M  & 16.58 \\ % luo2018neural, citing zoph2016neural  TODO check
            NAONet (f=36)           & 10.8M & 15.67  \\% luo2018neural
            AmoebaNet-B             & 34.9M & 15.80  \\% luo2018neural [38]
            NASNet-A(7@2304)        & 32.6M & 16.03  \\% luo2018neural, citing zoph2016neural  
            NAONet (f=128)          & 128M  & 14.75  \\% luo2018neural
            \midrule
            AdaNAS                    &  33M & $\mathbf{14.58 \pm 0.05}$ \\
            \bottomrule
        \end{tabular}
    \end{sc}\end{small}\end{center}
    \vskip -0.1in
\end{table}

}

    On \texttt{CIFAR-10} our results is surpassed only by NAONET with $4$ times more parameters and AmoebaNet, which used a different search space to discover its cell.
    
    On \texttt{CIFAR-100} our method's results surpasses the previous state-of-the-art.

    \subsection{Search space}
    
    As the dynamic generator explores different configurations of the NASNet architecture space, we observe an interesting behaviour between \texttt{CIFAR-10} and \texttt{CIFAR-100}.
    We observe (see Figure~\ref{img:experimentsearch}) that our algorithm consistently chose much wider networks for \texttt{CIFAR-100} than it did for \texttt{CIFAR-10}.
    The last architecture selected on \texttt{CIFAR-100} was NASNet-A(1@2640) ($8M$ parameters). 
    The last architecture selected on \texttt{CIFAR-10} was NASNet-A(6@1440) ($11M$ parameters).
    
    This is a valuable insight that even similar dataset may require very different architectures to perform well.
    This also means that eliminating the human bias even after NAS finds the building block is an important contribution.
    
    \begin{figure}[t]
    \vskip 0.2in
    \begin{center}
    \centerline{\includegraphics[width=\columnwidth]{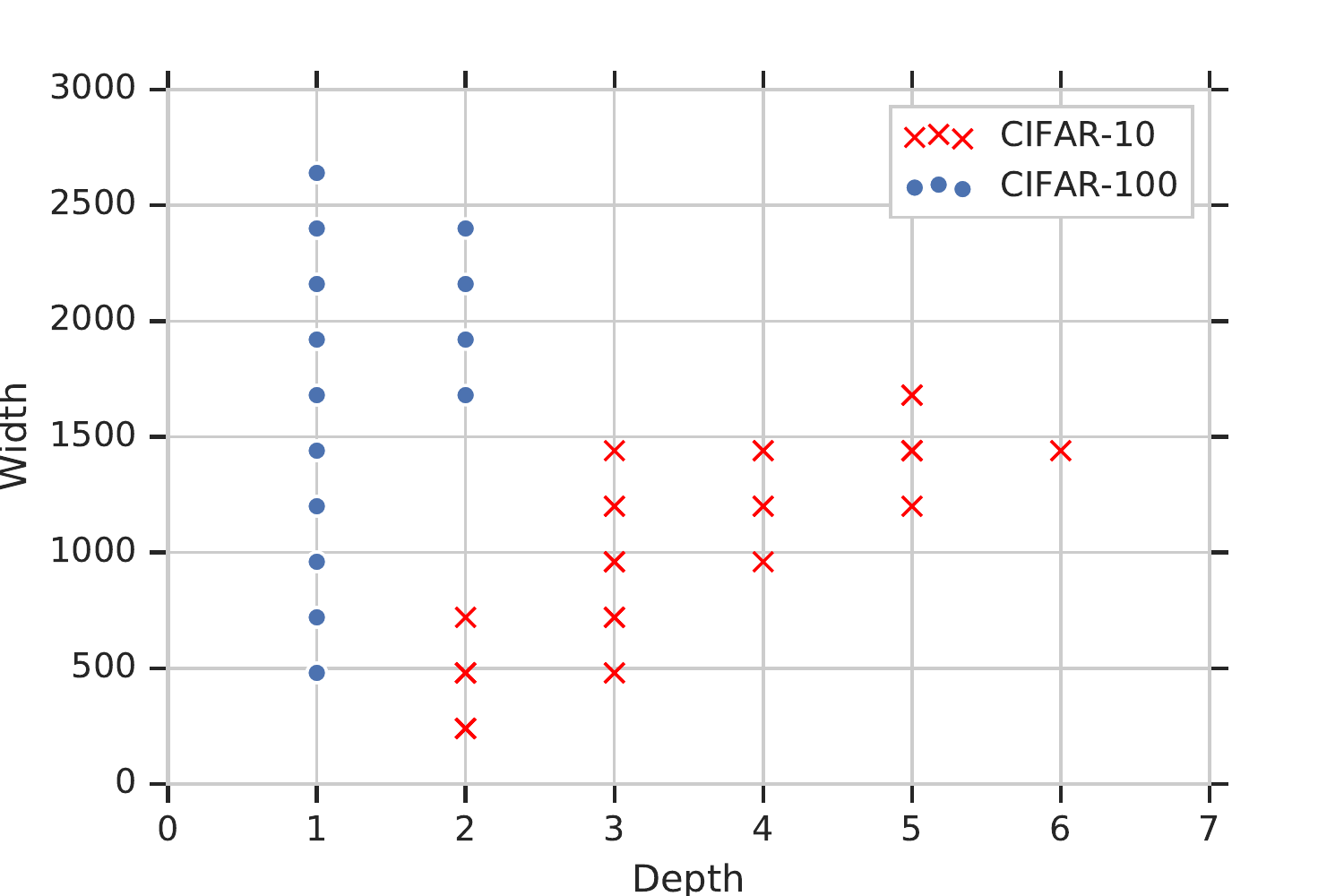}}
    \caption{
        Difference between explored architectures on \texttt{CIFAR-10} and \texttt{CIFAR-100}. 
        Each point represents one subnetwork that was chosen across $3$ experiments. 
        Depth is the first NASNet parameter (number of stacked cells) and width is the second (number of convolution filters).}
    \label{img:experimentsearch}
    \end{center}
    \vskip -0.2in
    \end{figure}

\section{Discussion} %
\label{sec:discussion}

    We have shown that our AdaNAS algorithm consistently improves the performance of NASNet models by learning an ensemble of smaller models with minimal hyperparameter tuning.
    We explored two methodologies for generating candidate subnetworks for the ensemble, and observed positive results in both cases. 
    Interestingly, a simple ensemble of identical architectures trained independently with a uniform averaged output already performs better than the baseline single large model. 
    The benefit is that these models can be trained in parallel and ensembled together in a single step, allowing for a minimal wall-clock time from training start to serving.
    
    Conversely, our adaptive methods show performance gains for applications where we can afford to train ensemble sequentially. 
    For one, we were able to achieve near state-of-the-art results by using a combination of learning mixture weights and applying Adaptive Knowledge Distillation. 
    We show that it is possible to make the ensembling process even more automatic by using a dynamic generator, that can produce models with minimal human knowledge and still outperform architectures designed by experts.

    Finally, as specialized multi-core machine learning hardware such as GPUs and TPUs become more ubiquitous, we envision model-parallelism at serving time to become increasingly important. 
    Ensembles of neural networks provide a simple form of model-parallelism, while producing high-quality models given a fixed parameter budget.

\section*{Acknowledgements}
    We would like to thank Scott Yak, Eugen Hotaj, Vitaly Kuznetsov, Mehryar Mohri, and Corinna Cortes for their insights and advice during this project. 
    We also extend a special thanks to our collaborators, residents and interns Gus Kristiansen, Galen Chuang, Ghassen Jerfel, Ben Adlam and the many others at Google and beyond who helped us shape AdaNet \cite{weill2018adanet}.

}

\bibliography{main}
\bibliographystyle{icml2019}

\end{document}